%% file: main.tex
\documentclass[conference]{IEEEtran}
\input{packages}
\IEEEoverridecommandlockouts
\def\BibTeX{{\rm B\kern-.05em{\sc i\kern-.025em b}\kern-.08em
    T\kern-.1667em\lower.7ex\hbox{E}\kern-.125emX}}
    
\include{commands}

\begin{document}

\title{Face Beneath the Ink: Synthetic Data and Tattoo Removal with Application to Face Recognition \\
{\large Mathias Ibsen, Christian Rathgeb, Pawel Drozdowski, Christoph Busch }
\thanks{The authors are with the Biometrics and Internet Security Research Group at Hochschule Darmstadt, Germany\\
E-mail: mathias.ibsen@h-da.de}
}
\author{}
\maketitle
\thispagestyle{plain}
\pagestyle{plain}

\begin{abstract}
Systems that analyse faces have seen significant improvements in recent years and are today used in numerous application scenarios. However, these systems have been found to be negatively affected by facial alterations such as tattoos. To better understand and mitigate the effect of facial tattoos in facial analysis systems, large datasets of images of individuals with and without tattoos are needed. To this end, we propose a generator for automatically adding realistic tattoos to facial images. Moreover, we demonstrate the feasibility of the generation by using a deep learning-based model for removing tattoos from face images. The experimental results show that it is possible to remove facial tattoos from real images without degrading the quality of the image. Additionally, we show that it is possible to improve face recognition accuracy by using the proposed deep learning-based tattoo removal before extracting and comparing facial features. 
\end{abstract}

\input{sections/introduction}
\input{sections/related_work}

\input{sections/generation}
\input{sections/tattoo_removal}
\input{sections/results}
\input{sections/conclusion}
\input{sections/acknowledgement}

\bibliographystyle{IEEEtran.bst}
\bibliography{bibli}
\end{document}

%% file: packages.tex
\usepackage{subcaption}
\usepackage{float}
\usepackage{multirow, multicol}
\usepackage{graphicx}
\usepackage{hhline}
\usepackage{amsmath}
\usepackage{array, makecell}%
\usepackage{bm}
\usepackage{amsbsy}
\usepackage{booktabs}
\usepackage{xspace}
\usepackage{xcolor}
\usepackage{newfloat}
\usepackage[export]{adjustbox}
\usepackage{enumitem}
\usepackage{url}

%% file: commands.tex
\def\eg{\textit{e.g.}\@\xspace} 
\def\ie{\textit{i.e.}\@\xspace}

\def\wrt{w.r.t.\@\xspace}
\def\etal{\textit{et al.}\@\xspace}

%% file: sections/introduction.tex
\section{Introduction} 
Facial analysis systems are deployed in various applications ranging from medical analysis to border control. Such facial analysis systems are known to be negatively affected by facial occlusions \cite{Zeng-SurveyOfFRUnderOcclusio-IET-2021}. A specific kind of facial alteration that partially occludes a face is a face tattoo. Facial tattoos have become more appealing recently and have been described as a mainstream trend in several major newspapers~\cite{nyt_face_mainstream, standard_face_mainstream}. Ensuring inclusiveness and accessibility for all individuals, independent of physical appearance, is imperative in developing fair facial analysis systems. In this regard, facial tattoos are especially challenging, as they cause permanent alterations where ink is induced into the dermis layer of the skin. For instance, Ibsen \etal investigated in \cite{Ibsen-ImpactFacialTattoosPaintingsFaceRecognitionSystems-BMT-2021} the impact of facial tattoos and paintings on state-of-the-art face recognition systems. The authors showed that tattoos might impair the recognition accuracy and thus the security of such a facial analysis system.

\begin{figure}
    \centering
    \includegraphics[width=0.75\linewidth]{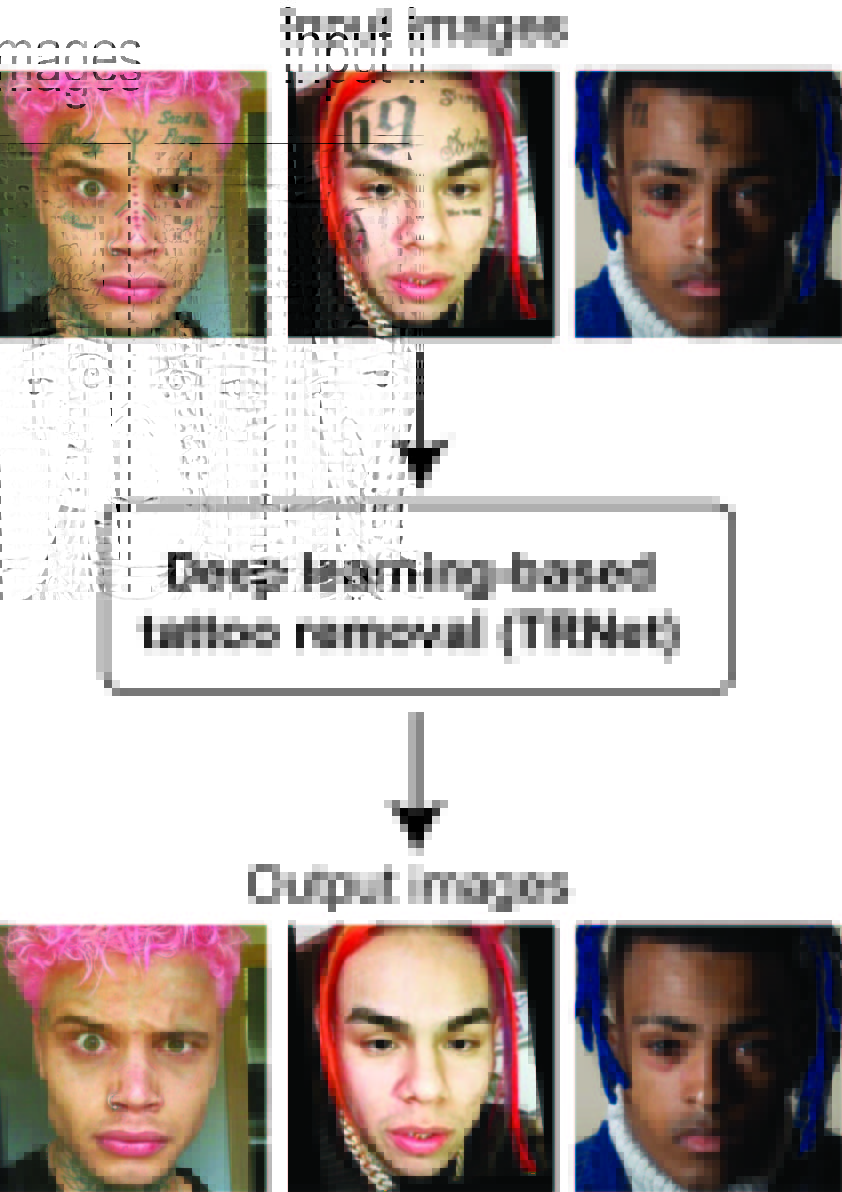}\vspace{0.1cm}
    \caption{Examples of using deep learning-based tattoo removal.}
    \label{fig:tattoo_removal_examples}
\end{figure}

In coherence with the findings in~\cite{Ibsen-ImpactFacialTattoosPaintingsFaceRecognitionSystems-BMT-2021}, it is of interest to make facial analysis systems more robust towards facial tattoos. One way to do this is face completion, where missing or occluded parts of a face are reconstructed; such approaches have, for instance, shown to improve face recognition performance for some occlusions \cite{Mathai-DoesGenerativeFaceCompletionHelpFR-ICB-2019}. An additional benefit of using face completion over approaches like occlusion-aware face recognition is the potential to use the reconstructed facial image for other purposes, e.g. visualising how a face might look without the occlusion or preventing that tattoos are used for recognition purposes; which is something that raises ethical issues as discussed in \cite{Bacchini-ATattooIsNotAFaceEthicalIssues-2017}.

However, one major problem with face completion for tattoo removal is the lack of sufficient and high-quality training data, as no extensive database of facial tattoos is currently available. 

The main focus of this work is, therefore, two-fold. First, we propose a method for synthetically adding tattoos to facial images, which we use to create a large database of facial images with tattoos. The proposed method uses face detection and landmark localisation to divide the face into regions, whereafter suitable placements of tattoos are found. Subsequently, we approximate depth and construct depth and cut-out maps which are used to realistically blend tattoos onto a face. It has recently been shown that synthetic data can be beneficial for face analysis tasks and be a good alternative to real data \cite{Wood-FakeItTillYouMakeIt-ICCV-2021}. Secondly, we show the usefulness of our synthetic data by training a deep learning-based model for tattoo removal (as illustrated in Fig.~\ref{fig:tattoo_removal_examples}) and evaluate the impact of removing facial tattoos on a state-of-the-art face recognition system using a database comprising real facial images with tattoos.  \par The approach for synthetically adding tattoos to a facial image in a fully automated way is, to the authors' best knowledge, the first of its kind. The proposed generator can be used to create large databases which can be used in related fields such as tattoo detection or studying the effects of tattoos on human perception. Additionally, we are the first to measure the effect of removing facial tattoos on face recognition systems. The code for synthetically adding tattoos to face images will be made publicly available\footnote{Generation code will be released upon acceptance of this manuscript}.

In summary, this work makes the following contributions:
\begin{itemize}
    \item A novel algorithm for synthetically adding facial tattoos to face images.
    \item An algorithm for removing tattoos from facial images trained on only facial images with synthetically added tattoos. We refer to this algorithm as TRNet.
    \item An experimental analysis of the quality of the tattoo removal.
    \item Showcasing the application of tattoo removal in a face recognition system by conducting an experimental analysis on the effect of removing facial tattoos on a face recognition system.
\end{itemize}

The outline of the remaining article is as follows: Sect.~\ref{sec:related_work} describes prominent related works, Sect.~\ref{sec:synth_data_gen} describes an automated approach for synthetically blending tattoos to facial images, which is used in Sect.~\ref{sec:generated_database} to generate a database of facial images with tattoos. Sect.~\ref{sec:tattoo_removal} and \ref{sec:experiments} show the feasibility of the synthetic generation by training a deep learning-based model for tattoo removal and evaluating if it can improve biometric recognition performance, respectively. Finally, Sect.~\ref{sec:conclusion} provides a summary of this work. 

%% file: sections/related_work.tex
\section{Related Work}
\label{sec:related_work}
The following subsections summarise related works \wrt synthetic data generation for facial analysis (Sect.~\ref{sec:subsec_synth_data_gen}),  facial alterations (Sect.~\ref{sec:facial_alterations}), and facial completion (Sect.~\ref{sec:facial_completion}).

\subsection{Synthetic Data Generation for Face Analysis}
\label{sec:subsec_synth_data_gen}
Synthetically generated data has seen many application scenarios in face analysis, most notably for addressing the lack of training data. Synthetic data has especially become relevant with the recent advances in deep learning-based algorithms, which usually require a large amount of training data. Privacy regulations, \eg the European General Data Protection Regulation~\cite{EU-Regulation-2016-679-on-DataPrivacy-160427}, make sharing and distributing large-scale face databases impracticable as face images are classified as a special category of personal data when used for biometric identification. As an alternative, researchers have explored the use of synthetic data. The generation of realistic-looking synthetic face data has especially become feasible with the recent advances in Generative Adversarial Networks (GANs), first proposed by Goodfellow \etal in~\cite{Goodfellow-GenerativeAdversarialNetworks-2014}. Prominent work in this field includes StyleGAN, which was first introduced in \cite{Karras-StyleGAN-CVPR-2019} by Karras \etal and showed, at the time, state-of-the-art performance for synthesising facial images. Since the original work, two improved versions of StyleGAN have been proposed~\cite{Karras-StyleGAN2-CVPR-2021, Karras-StyleGAN3-NeurIPS-2021}. Much current research in this area focuses on GAN-inversion, where existing face images are encoded into the latent space of a generator. Thereafter, the resulting latent code can be shifted in the latent space, whereby the inverted image of the shifted vector results in an alteration of the original image. The technique can, for instance, be used for face age progression~\cite{Grimmer-DeepFaceAgeProgressionSurvey-IEEEAccess-2021}. In addition to the face, some research has also been conducted for other biometric modalities, \eg fingerprint~\cite{Cappelli-SFinGE-2004, Priesnitz-SynCoLFinGer-arxiv-2021, Wyzykowski-L3SyntheticFingerprintGeneration-ICPR-2021} and iris~\cite{Drozdowski-SICGen-BIOSIG-2017, mitre_iris_generation}.

Little work has been conducted regarding synthetic data generation of facial images with tattoos. However, in \cite{Xu-PortraitPhotoToTattooTransformBasedOnDigitialTattooing-MultimediaToolsAndApplications-2021} the authors proposed a method for transforming digital portrait images into realistic-looking tattoos. In \cite{skindeep}, the author also shows examples of tattoo images added to facial and body images using an existing GAN for drawing art portraits; however, details about this approach are not scientifically documented.

\begin{figure*}[htbp!]
\centering
\includegraphics[width=0.7\linewidth]{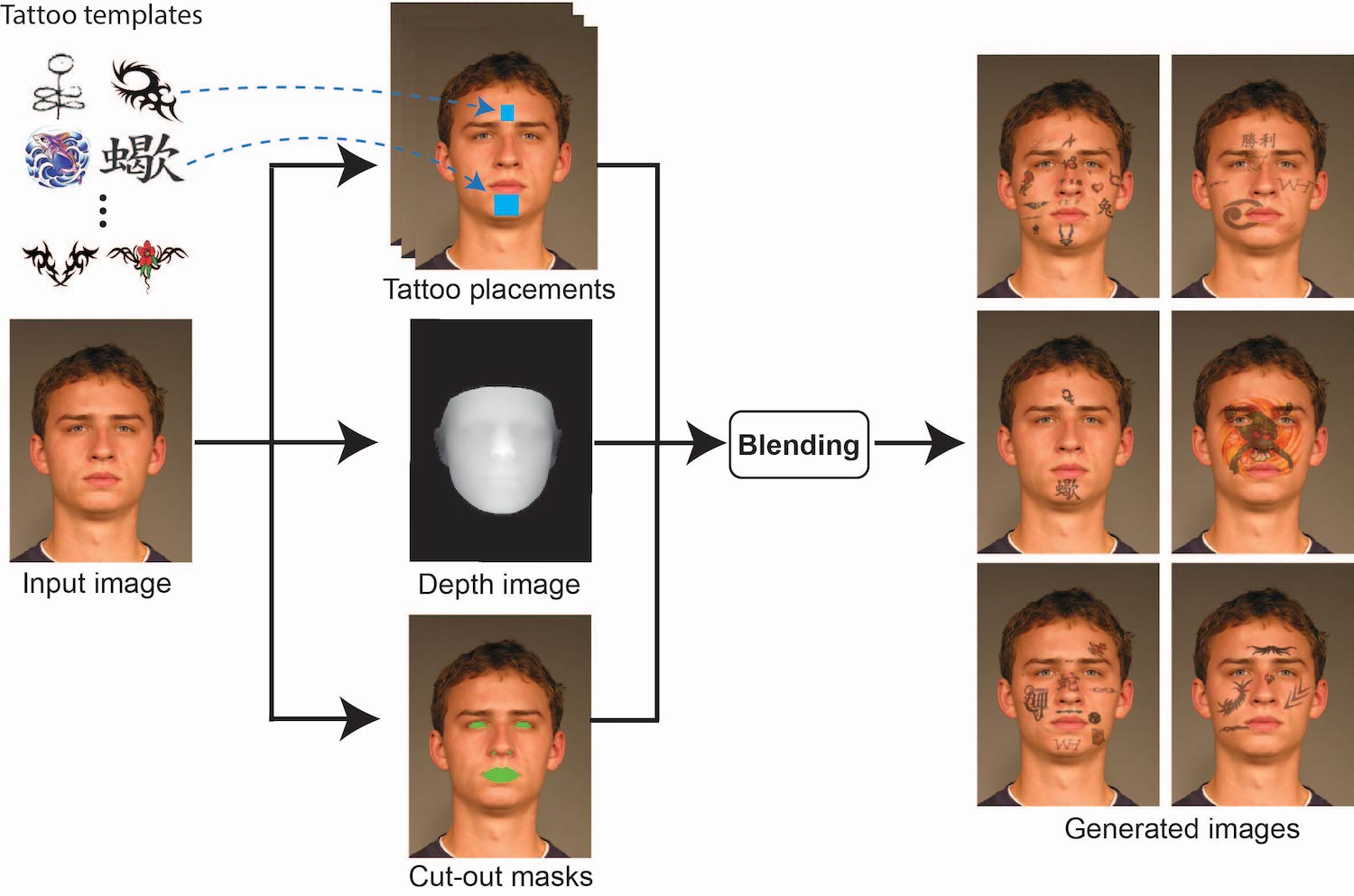}
\caption{Synthetic facial tattoo generation workflow.}
\label{fig:selfie_filter_creation}
\end{figure*}

\subsection{Facial Alterations}
\label{sec:facial_alterations}
Facial alterations can occur in either the physical or digital domain and cause permanent or temporary changes of a face. Several studies have explored the impact of physical and digital alterations on face recognition systems. In the physical domain, predominantly the effects of makeup and plastic surgery on face recognition have been studied \cite{Rathgeb-ImpactDetectionFacialBeautificationSurvey-ACCESS-2019}. In~\cite{Singh-PlasticSurgeryANewDimensionToFaceRecogniton-ieee} the authors collected a database of 900 individuals to analyse the effect of plastic surgery and found that the tested algorithms were unable to effectively account for the appearance changes caused by plastic surgery. More recently, Rathgeb \etal showed in~\cite{Rathgeb-PlasticSurgeryDeepFace-CVPRW-2020}, using a database of mostly ICAO-quality face images~\cite{ICAO-9303-p9-2015} captured before and after various types of facial plastic surgeries, that different tested state-of-the-art face recognition systems maintained almost perfect verification performance at an operationally relevant threshold corresponding to a False Match Rate (FMR) of $0.1\%$. Numerous works have addressed the impact of temporary alterations on face recognition systems. In~\cite{Dantcheva-CanFacialCosmeticsAffectTheMatchingAccuracyOfFaceRecognitionSystems}, Dantcheva \etal found that makeup can hinder reliable face recognition; similar conclusions were drawn by Wang \etal in~\cite{Wang-Recognizing-HumanFacesUnderDisguiseAndMakeup-IEEE-2016} where they investigated the impact of human faces under disguise and makeup. The previous work shows that makeup might be successfully used for identity concealment;  in~\cite{Chen-SpoofingFacesUsingMakeupAnInvestigativeStudy-IEEE-2017}, the authors additionally showed that makeup could also be used for presentation attacks with the goal of impersonating another identity. In~\cite{Rathgeb-MakeupAttackDetection-IWBF-2020}, the authors found that especially high-quality makeup-based presentation attacks can hamper the security of face recognition systems. In~\cite{Singh-RecognizingDisguisedFacesInTheWild-TBIOM-2019}, the authors found that disguised faces severely affect recognition performance, especially for occlusions near the periocular region. The database used by the authors includes different types of disguises, including facial paintings. Coherent with these findings, Ibsen \etal showed in \cite{Ibsen-ImpactFacialTattoosPaintingsFaceRecognitionSystems-BMT-2021} that facial tattoos and paintings can severely affect different modules of a face recognition system, including face detection as well as feature extraction and comparison.

Ferrara \etal were among the first to show that digital alterations can impair the security of face recognition systems. Especially notable is their work in~\cite{Ferrara-TheMagicPassport-IJCB-2014} where they showed the possibility of attacking face recognition systems using morphed images. Specifically, they showed that if a high-quality morphed image is infiltrated into a face recognition system (\eg stored in a passport), it is likely that individuals contributing to the morph are positively authenticated by the biometric system. Since then, there have been numerous works on face recognition systems under morphing attacks. For a comprehensive survey, the reader is referred to~\cite{Scherhag-MorphingSurvey-IEEE-2019}. Facial retouching is another area which has seen some attention in the research community. While some early works showed that face recognition can be significantly affected by retouching, Rathgeb \etal showed more recently that face recognition systems might be robust to slight alterations caused by retouching \cite{Rathgeb-PRNU-Retouching-Detection-BMT-2020}. Similar improvements have been shown for geometrical distortions, \eg stretching~\cite{Hedberg-EffectOfSampleStretchingOnFR-Biosig-2020}. A more recent threat that has arrived with the prevalence of deep-learning techniques is so-called DeepFakes~\cite{Verdoliva-MedianForensicsAndDeepFakeOverview-2020}, which can be used to spread misinformation and as such lead to a loss of trust in digital content. Many researchers are working on the detection or generation of deep learning-based alterations. Several arduous challenges and benchmarks have already been established, for instance, the recent \textit{Deepfake Detection Challenge}~\cite{Ferrer-DeepfakeDetectionChallengeResults-online-2020}  where the top model only achieved an accuracy of approximately $65\%$ on previously unseen data. Generation and detection of deep learning-based alterations are continuously evolving and remain a cat-and-mouse game; interested readers are referred to~\cite{TolosanaDeepFakesBeyondSurveyOfFaceManipulationFaceDetection2020} for a comprehensive survey. 

\subsection{Facial Completion}
\label{sec:facial_completion}
Most methods for face completion\footnote{Also called face inpainting.} build upon deep learning-based algorithms, which are trained on paired images where each pair contains a non-occluded face and a corresponding occluded face. In \cite{Iizuka-GloballyAndLocallyConsistentImageCompletion-}, the authors proposed an approach for general image completion and showed its applicability for facial completion. In this work, the authors leveraged a fully convolutional neural network trained with global and local context discriminators. Similar work was done in \cite{Li-GenerativeFaceCompletion-CVPR-2017} where the authors occluded faces by adding random squares of noise pixels. Subsequently, they trained an autoencoder to reconstruct the occluded part of the face using global and local adversarial losses as well as a semantic parsing loss. Motivated by the prevalence of VR/AR displays which can hinder face-to-face communication, Zhao \etal \cite{Zhao-IdentityPreservingFaceCompletion-BMVC-2018} proposed a new generative architecture with an identity preserving loss. In \cite{Song-GemotryAwareFaceCompletionAndEditing-AAAI-2019}, Song \etal used landmark detection to estimate the geometry of a face and used it, together with the occluded face image, as input to an encoder-decoder architecture for reconstructing the occluded parts of the face. The proposed approach allows generating diverse results by altering the estimated facial geometry. More recently, Din \etal \cite{Din-GanBasedUnmaskingOfFace-IEEEAccess-2020} employed a GAN-based architecture for the unmasking of masked facial images. The proposed architecture consists of two stages where the first stage detects the masked area of the face and creates a binary segmentation map. The segmentation map is then used in the second stage for facial completion using a GAN-based architecture with two discriminators where one focuses on the global structure and the other on the occluded parts of the face. In \cite{Mathai-DoesGenerativeFaceCompletionHelpFR-ICB-2019}, it was found that facial completion can improve face recognition performance.

%% file: sections/generation.tex
\section{Facial Tattoo Generator}
\label{sec:synth_data_gen}
To address the lack of existing databases of image pairs of individuals before and after they got facial tattoos, we propose an automated approach for synthetically adding facial tattoos to images. An overview of the proposed generation is depicted in Fig.~\ref{fig:selfie_filter_creation}. The process of synthetically adding tattoos to a facial image can be split into two main steps which are described in the following subsections: (1) finding the placement of tattoos in a face and (2) blending the tattoos to the face. 

\subsection{Placement of Tattoos}
\label{sec:placement}
\begin{figure}[!htbp]
    \centering
    \includegraphics[width=0.50\linewidth]{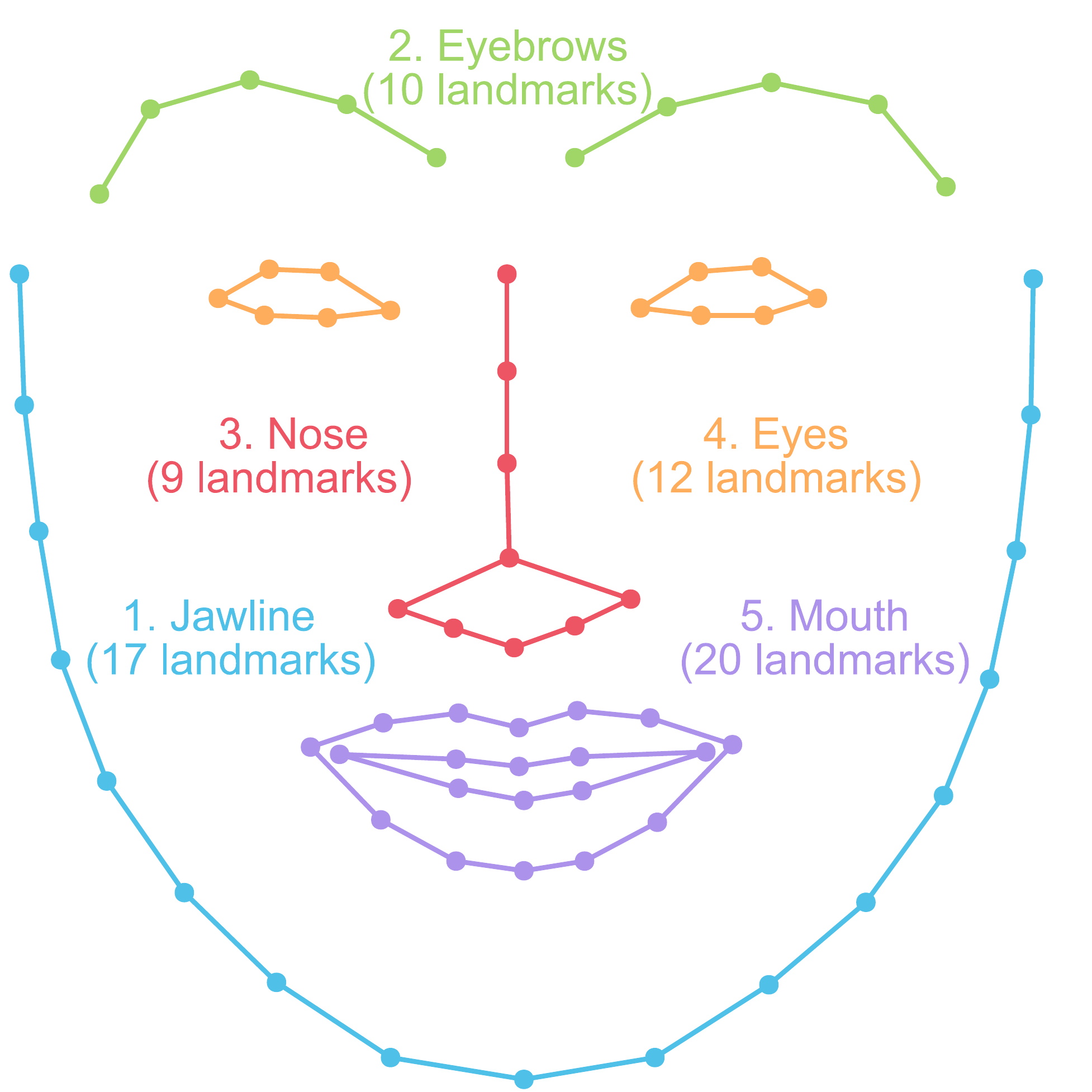}
    \caption{Facial landmarks detected by dlib.}
    \label{fig:dlib_landmarks}
\end{figure}

To find suitable placements of tattoos in a face, we start by localising the facial region and detecting landmarks of the face. To this end, we use \textit{dlib}~\cite{King-MachineLearningToolkit-2009} which returns a list of 68 landmarks as shown in Fig. \ref{fig:dlib_landmarks}. 

\begin{figure}[!htbp]
\begin{subfigure}[t]{0.3\linewidth}
    \centering
  \includegraphics[width=\textwidth]{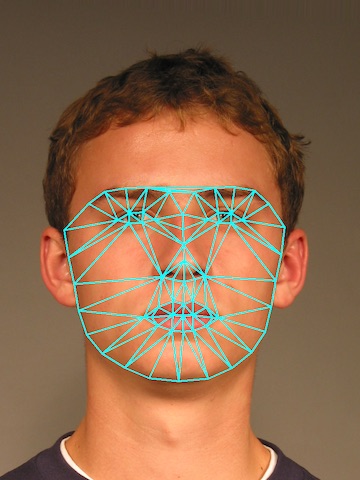}
        \caption{}
\end{subfigure}\hfil %
\begin{subfigure}[t]{0.3\linewidth}
    \centering
  \includegraphics[width=\textwidth]{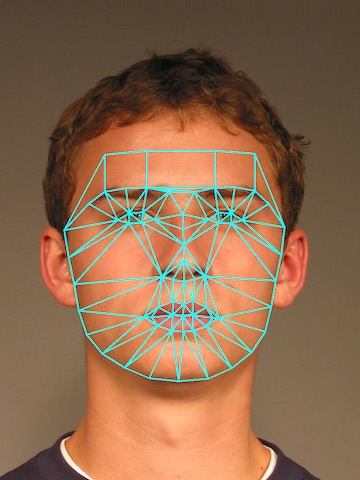}
        \caption{}
\end{subfigure}\hfil %
\begin{subfigure}[t]{0.3\linewidth}
    \centering
  \includegraphics[width=\textwidth]{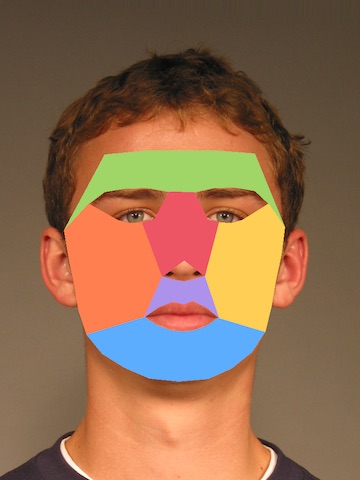}
  \caption{}
  \label{fig:predefined_regions}.
\end{subfigure}
    \caption{(a) Division of a facial image into regions from landmarks, (b) extended to the forehead, and (c) division into six pre-defined regions.}
\label{fig:initial_regions}
\end{figure}

Thereafter, the landmarks are used to divide the face into small regions of triangles by performing a fixed Delaunay triangulation. The regions are then extended to the forehead by using the length of the nose as an estimate. Each region now constitutes a possible placement of a tattoo; however, such a division is inadequate for the placement of larger tattoos. Therefore, the face is divided into six larger regions. The division of the face into large and small regions gives high controllability in the data generation. As indicated, some regions are excluded, \ie regions around the nostrils, mouth and nose. These regions are estimated based on the detected landmarks. The division of a face into regions is illustrated in Fig.~\ref{fig:initial_regions}. The regions make it possible to avoid placing tattoos in heavily bearded areas or on top of glasses if such information is available about the facial images during the generation phase. In our work, we do not use beard or glass detectors; however, for some of the images information about beard or glasses is available which we use to avoid placing tattoos in the affected regions.

\begin{figure}[!htbp]
\begin{subfigure}[t]{0.3\linewidth}
    \centering
  \includegraphics[width=\textwidth]{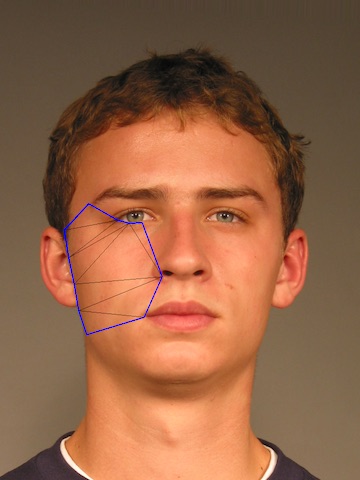}
  \caption{Selected region}
\end{subfigure}\hfil %
\begin{subfigure}[t]{0.3\linewidth}
    \centering
  \includegraphics[width=\textwidth]{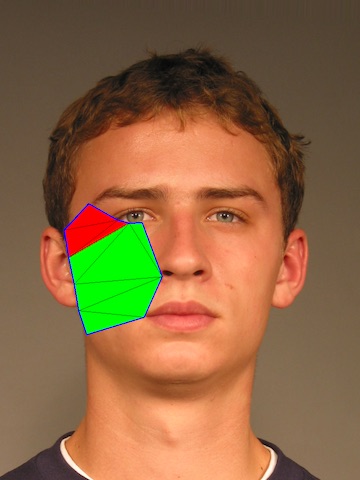}
  \caption{Find a subset of the region not occupied (the green area).}
\end{subfigure}\hfil %
\begin{subfigure}[t]{0.3\linewidth}
    \centering
  \includegraphics[width=\textwidth]{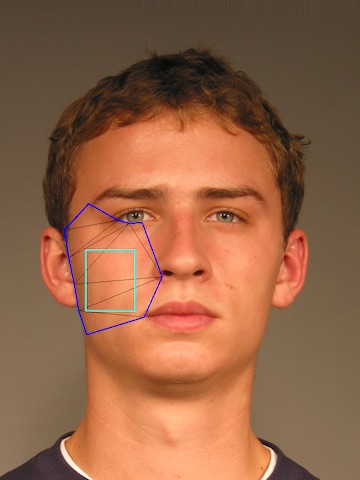}
  \caption{Find a placement for the tattoo}
\end{subfigure}
\caption{Illustration showing an example of how a placement of a tattoo in a region can be found. The red area in (b) illustrates that there might be some areas within a selected region where a tattoo cannot be placed, \eg if the area is reserved for another tattoo.}
\label{fig:finding_placement_ex}
\end{figure}

A tattoo can now be placed in one of the six predefined regions, or the regions can be further combined to allow placing the tattoos in larger areas of the face. A combined region is simply a new region consisting of several smaller regions. The exact placement of a tattoo within a region depends on a pre-selected generation strategy. The generation strategy determines (1) possible regions where a tattoo can be placed, (2) the selection of tattoos, and (3) the size and placement of a tattoo within a region. An example is illustrated in  Fig.~\ref{fig:finding_placement_ex} where one of the cheeks is selected as a possible region, whereafter the largest unoccupied subset within that region is found. Thereafter, the tattoo is placed by estimating its largest possible placement within the selected subset without altering the original aspect ratio of the tattoo. In this work, we use a database comprising more than 600 distinct tattoo templates, which mainly consist of real tattoo designs collected from acquired tattoo books. The selection of which tattoos to place depends on the generation strategies, which are further described in Sect.~\ref{sec:generation_strategy}.

\subsection{Blending}
\label{sec:blending}
To blend the tattoos to faces, various image manipulations are performed.\par Given a facial image and placement of tattoos (see Sect.~\ref{sec:placement}); each tattoo is placed and overlayed on the facial image by multiplying the tattoo layer with the facial image. Afterwards, the tattoo is displaced to match the contours of the face using displacement mapping. Areas of the tattoo which have been displaced outside the face or inside the mouth, nostrils and eyes are cut out. This is done by using cut-out maps (see Fig.~\ref{fig:selfie_filter_creation}), which are calculated from the landmarks detected by dlib in the placement phase. Lastly, the tattoo is made more realistic by colour adjustment, Gaussian blurring, and lowering the opacity of the tattoo. 

As previously stated, displacement mapping is used for mapping tattoos to the contours of a face. It is a technique which utilises depth information of texture maps to alter the positions of pixels according to the depth information in the provided map. Contrary to other approaches, such as bump mapping, it alters the source image by displacing pixels. In displacement mapping, a map $M$ containing values in the range 0-255 is used to displace pixels in a source image $I$. In general, a specific pixel, $I(x,y)$, is displaced in one direction if $M(x,y)$ is less than the theoretical average pixel value of the map (127.5); else it is displaced in the opposite direction. For the displacement technique used in this work, a pixel in the source image is displaced both vertically and horizontally. 

More specifically, let $c$ be a coefficient, let $(x,y) \in I$, and let $(x,y) \in M$. The distance for displacing a pixel, $I(x,y)$, in the vertical and horizontal direction is then:

$$D(x,y) = c \cdot \frac{M(x,y) - 127.5}{127.5} $$

\begin{figure}[!htb]
    \begin{subfigure}[t]{0.5\linewidth}
    \centering
        \includegraphics[width=0.6\textwidth]{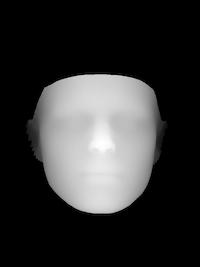}
        \caption{Depth image generated by PRNet}
        \label{fig:depthmap}
    \end{subfigure}%
    \begin{subfigure}[t]{0.5\linewidth}
    \centering
        \includegraphics[width=0.6\textwidth]{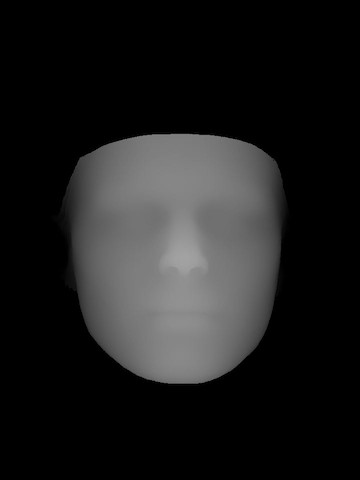}
        \caption{Transformed depth image}
        \label{fig:depthmap_transform}
    \end{subfigure}
    \caption{Example of (a) a depth map generated from a facial image using PRNet  and (b) after it has been transformed.}
    \label{fig:depmap_transformation_compare}
\end{figure}

PRNet \cite{feng2018prn} is used to generate the depth maps used in this work. PRNet is capable of performing 3D face reconstruction from 2D face images, and as such, it can also approximate depth maps from 2D facial images. An example of a depth map generated using PRNet is shown in Fig. \ref{fig:depthmap}.

\begin{figure}[!htbp]
    \begin{subfigure}{0.48\linewidth}
    \centering
        \includegraphics[width=0.6\textwidth]{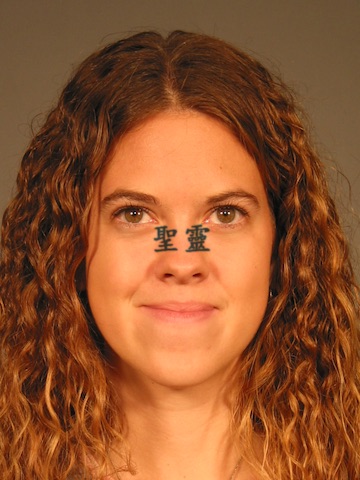}
        \caption{not displaced}
    \end{subfigure}
    \begin{subfigure}{0.48\linewidth}
    \centering
        \includegraphics[width=0.6\textwidth]{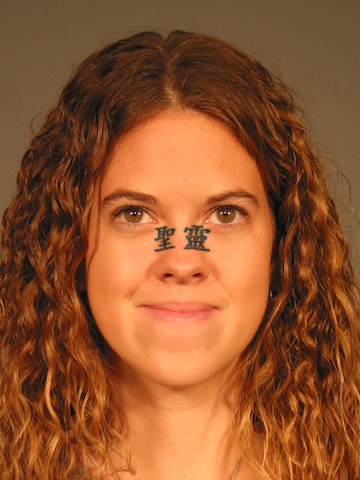}
        \caption{displaced}
    \end{subfigure}
    \caption{Facial images with tattoos (a) before and (b) after applying the displacement technique. For (b) the tattoo is bended around the anticipated 3D shape of the nose. Best viewed in electronic format (zoomed in).}
    \label{fig:displacements_examples}
\end{figure}

As seen in Fig. \ref{fig:depthmap}, the pixel values in the face region are rather bright, and there is little contrast. The small contrast between the pixel values and the high offset from the theoretical average pixel value implies that the depth map will not work very well as tattoos will be displaced too much in certain regions and too little in others. Therefore, to make the displacement more realistic, the depth map generated by PRNet is transformed by increasing the contrast and lowering the brightness of the map. Fig. \ref{fig:depthmap_transform} shows an example of a transformed depth map, and as it can be seen, the pixel values are much closer to the theoretical average value than the unaltered map, while the contrast around the nose, eyes and mouth are still high. Fig. \ref{fig:displacements_examples} shows an example where two facial tattoos are displaced to match the contours of a face.

\begin{figure}[!htb]
    \centering %
\begin{subfigure}[t]{0.3\linewidth}
  \includegraphics[width=\linewidth]{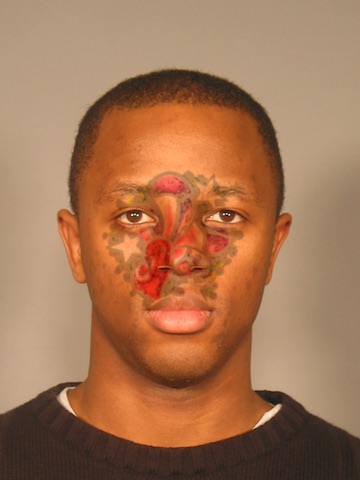}
\end{subfigure}\hfil %
\begin{subfigure}[t]{0.3\linewidth}
  \includegraphics[width=\linewidth]{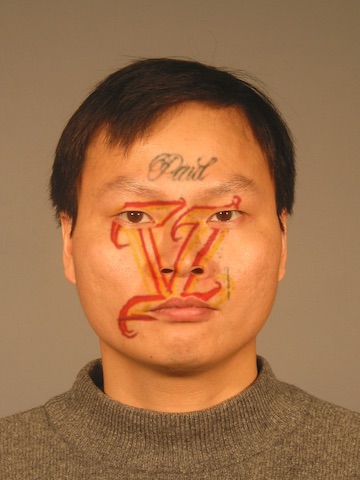}
\end{subfigure}\hfil %
\begin{subfigure}[t]{0.3\linewidth}
  \includegraphics[width=\linewidth]{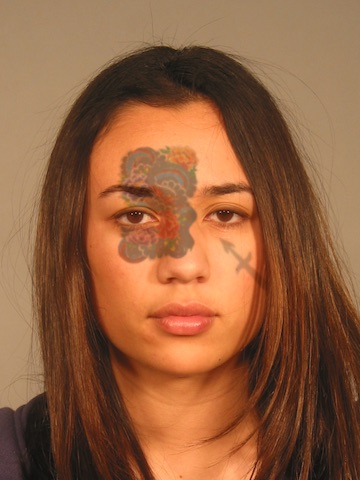}
\end{subfigure}
\caption{Examples of facial images where parts of one or more tattoos have been cut out.}
\label{fig:tattoo_cut_out}
\end{figure}

\begin{figure}[!htbp]
    \centering %
\begin{subfigure}[t]{0.3\linewidth}
  \includegraphics[width=\linewidth]{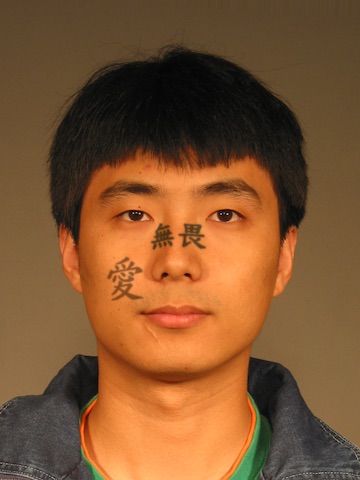}
\end{subfigure}\hfil %
\begin{subfigure}[t]{0.3\linewidth}
  \includegraphics[width=\linewidth]{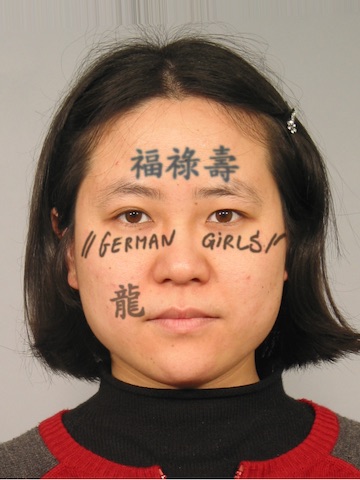}
\end{subfigure}\hfil %
\begin{subfigure}[t]{0.3\linewidth}
  \includegraphics[width=\linewidth]{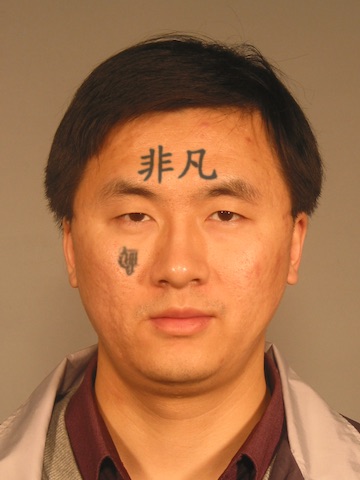}
\end{subfigure}
    \caption{Examples of black tattoos blended to facial images.}
    \label{fig:black_tattoos}
\end{figure}

Black ink tends to change in colour slightly over time due to the pigment used in black ink. Therefore, for colour adjustment, all pixels of a tattoo which is similar to pure black are selected and changed to simulate different colours of grey, green, and blue, which causes black tattoos to appear different for different facial images. The colour adjustments of black pixels are determined per tattoo, and as such slight variations can occur between different tattoos in the same facial image. Examples are given in Fig. \ref{fig:black_tattoos}. 

\subsection{Generation Strategies}
\label{sec:generation_strategy}
\begin{figure}[!htbp]
\centering
\begin{subfigure}{0.3\linewidth}
  \centering
  \includegraphics[width=\linewidth]{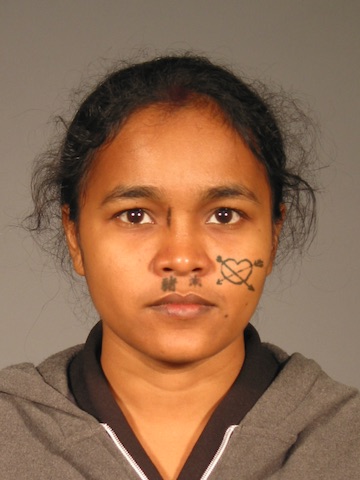}  
  \caption{} 
    \label{fig:generation_types_cov5}
\end{subfigure}\hfil 
\begin{subfigure}{0.3\linewidth}
  \centering
  \includegraphics[width=\linewidth]{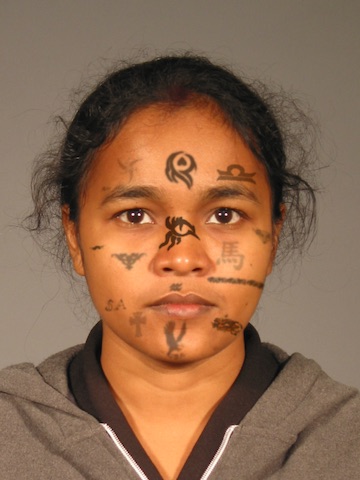}  
  \caption{}
\end{subfigure}\hfil 
\begin{subfigure}{0.3\linewidth}
  \centering
  \includegraphics[width=\linewidth]{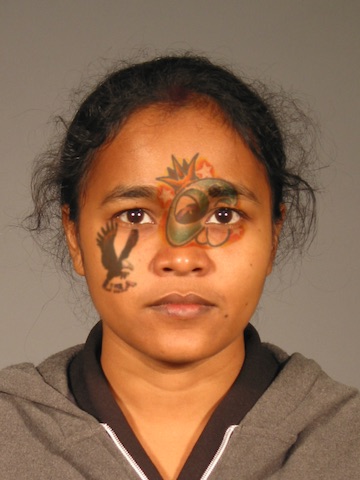}  
  \caption{}
  \label{fig:generation_types_cov25}
\end{subfigure}\hfil 
\bigskip
\begin{subfigure}{0.3\linewidth}
  \centering
  \includegraphics[width=\linewidth]{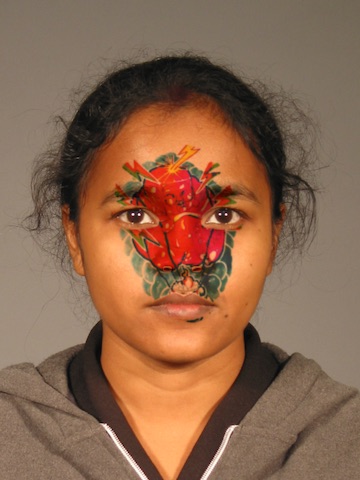}  
  \caption{}
    \label{fig:generation_types_extreme}
\end{subfigure}\hfil 
\begin{subfigure}{0.3\linewidth}
  \centering
  \includegraphics[width=\linewidth]{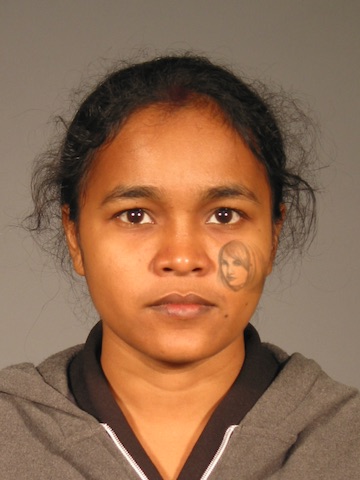}  
  \caption{}
  \label{fig:generation_types_portrait}
\end{subfigure}\hfil 
\begin{subfigure}{0.3\linewidth}
  \centering
  \includegraphics[width=\linewidth]{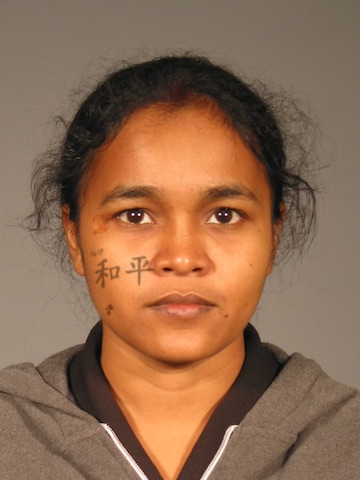}  
  \caption{}
\label{fig:generation_types_leftcheek}
\end{subfigure}\hfil 
\caption{Examples for different types of tattooed face that can be generated: (a) 5\%, (b) 15\%,  (c) 25\% coverage, (d) entire face, (e) single tattoo, and (f) specific region.}
\label{fig:generation_types}
\end{figure}

By varying how tattoos are selected and placed (Sect.~\ref{sec:placement}), many different types of facial images with tattoos can be generated. For the database used in this work, we employed two different strategies. In the first strategy, a desired coverage percent of tattoos on a face is randomly chosen from a specified range. Subsequently, tattoos are arbitrarily selected and placed on facial regions until the resulting coverage approximates the desired coverage. The coverage of a tattoo on a face is calculated based on the total amount of facial regions to which tattoos can be placed (see Fig.~\ref{fig:predefined_regions}) and the number of non-transparent pixels in the placed tattoos. In the second strategy, a specific region is always selected. Using the first strategy, it is possible to create databases where tattoos are placed arbitrarily until a selected coverage percent has been reached (see Fig.~\ref{fig:generation_types_cov5}-\ref{fig:generation_types_cov25}). Using the latter approach allows for more controlled placement of tattoos, \eg placing tattoos in the entire face region (Fig.~\ref{fig:generation_types_extreme}) or in a specific region (Fig.~\ref{fig:generation_types_portrait}-\ref{fig:generation_types_leftcheek}).

%% file: sections/tattoo_removal.tex
\section{Synthetic Tattoo Database}
\label{sec:generated_database}
This section describes the generation of a large database of facial images with tattoos. The database is used in section \ref{sec:tattoo_removal} to train deep learning-based models for removing tattoos from facial images. To generate the synthetic tattoo database, subsets of original images from the FERET~\cite{Phillips-FERET-1998}, FRGCv2~\cite{Phillips-FRGC-2005}, and CelebA~\cite{liu2015faceattributes} datasets were used. An overview of the generated database is given in Table~\ref{tab:generated_database}. For the FERET and FRGCv2 datasets, different generation strategies were used, including facial images where tattoos have been placed randomly, with specific coverage ranging from 5$\%$ to 25$\%$ as well as placement of single tattoos. For the single tattoos, we generated two versions: one version where the tattoo is placed in the entire facial region and another where portrait tattoos are blended to a random region in the face. For the CelebA database, which is more uncontrolled, facial tattoos were placed randomly. To simulate varying image qualities, a data augmentation was performed by randomly applying differing degrees of JPEG compression or Gaussian blur to all the images. Tattoo images and corresponding original (bona fide) images were paired such that similar augmentation was applied to corresponding images. 

\begin{table}[!htbp]
    \centering
\caption{Overview of the generated database (before augmentation).}
    \begin{tabular}{@{\extracolsep{2pt}}lrrr@{}} \toprule 
     \multirow{2.5}{*}{\textbf{Database}} & \multirow{2.5}{*}{\textbf{Subjects}} &
      \multicolumn{2}{c}{\textbf{Images}}  \\  \cmidrule{3-4}
     & & \textbf{Bona fide} & \textbf{Tattooed} \\ \midrule
    FERET  & 529 & 621  & 6,743 \\
    FRGCv2 & 533 & 1,436 & 16,209 \\
    CelebA & 6,872 & 6,872 & 6,872 \\
    \bottomrule
    \end{tabular}
\label{tab:generated_database}
\end{table}

\begin{figure*}
    \centering
    \includegraphics[width=0.7\linewidth]{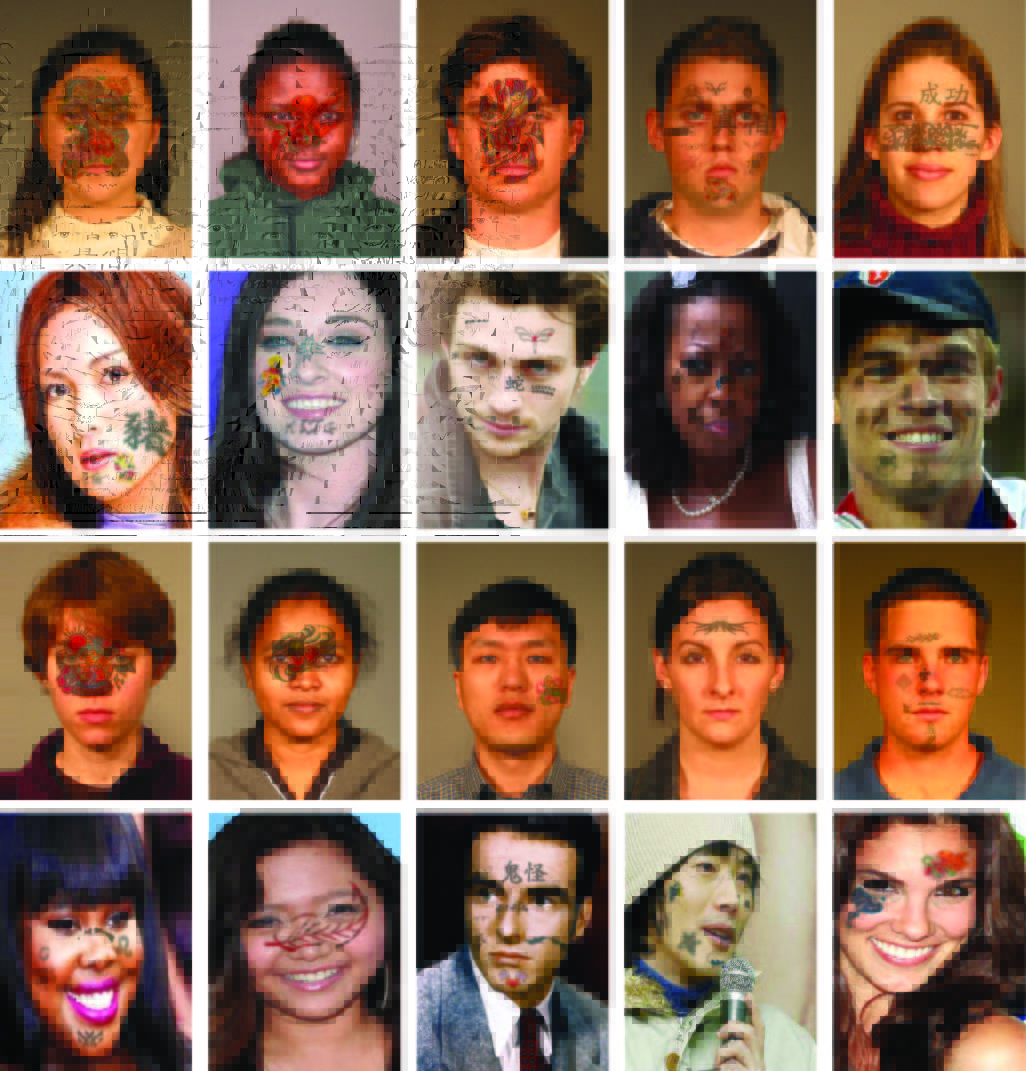}
    \caption{Examples of generated facial images with tattoos.}
    \label{fig:generated_examples}
\end{figure*}

Examples of images in the generated database are depicted in Fig.~\ref{fig:generated_examples}. 

\section{Tattoo Removal}
\label{sec:tattoo_removal}
To evaluate the realism of the proposed data generation and its use in real-world applications, two models are trained for the task of tattoo removal. Sect.~\ref{sec:tattoo_removal_models} briefly describes the different models used for removing tattoos. Sect.~\ref{sec:metrics} describes different metrics for evaluating the quality of the tattoo removal which is then evaluated in Sect.~\ref{sec:tattoo_removal_results}.

\subsection{Models}
\label{sec:tattoo_removal_models} 
Two different deep learning-based methods were trained for removing tattoos from facial images:
\begin{description}
\item[\textbf{pix2pix}] is a supervised conditional GAN for image-to-image translation \cite{isola2017image}. For the generator, a U-Net architecture is used, whereas the discriminator is based on a PatchGAN classifier which divides the image into $N \times N$ patches and discriminates between bona fide (\ie, real images) and fake images. 
\end{description}

\begin{figure*}[t]
\centering
\includegraphics[width=0.95\linewidth]{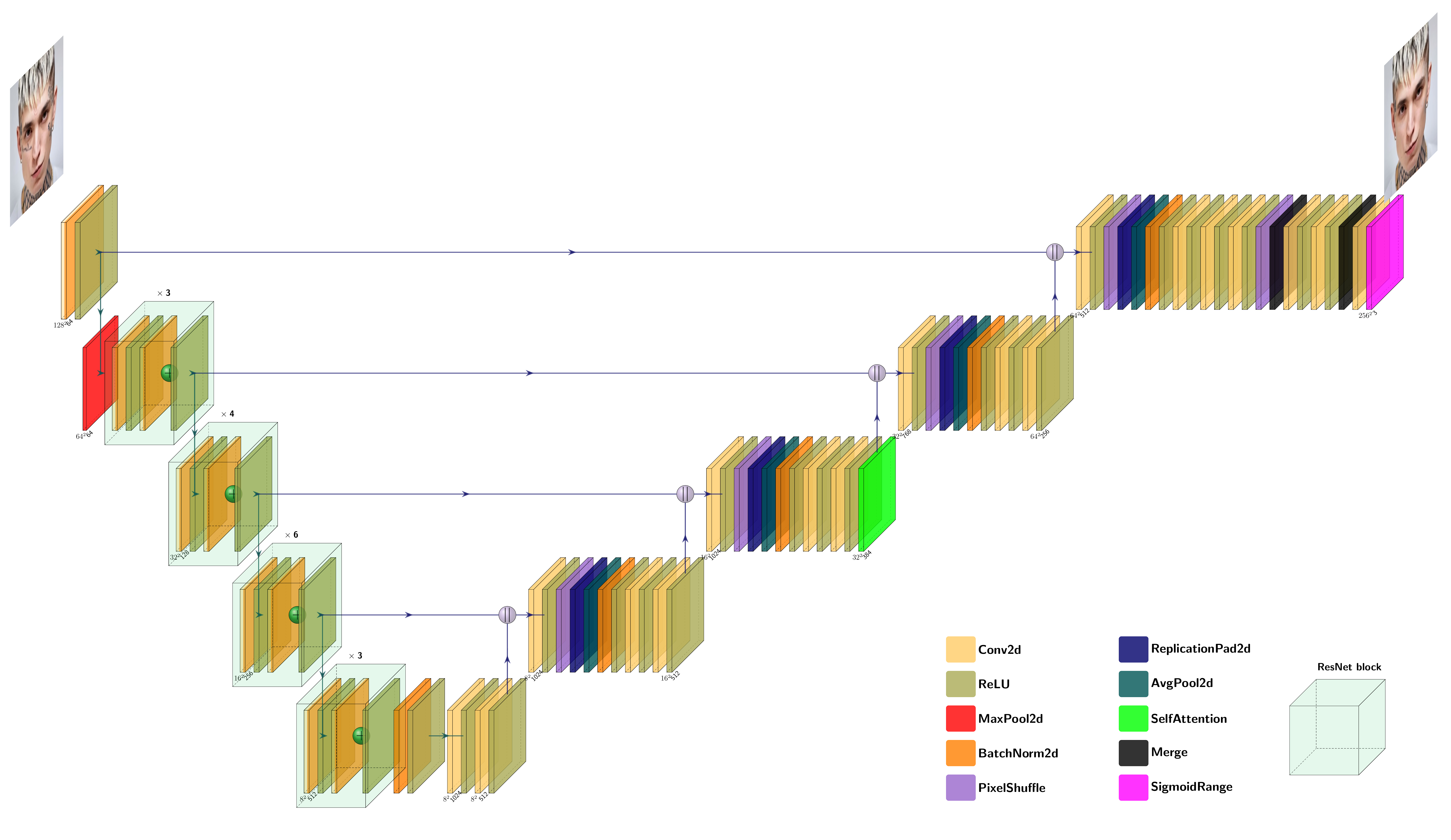}
\caption{Architecture of the tattoo removal network (TRNet).}
\label{fig:tnet_arch}
\vspace{-10pt}
\end{figure*}

\begin{description}
\item[\textbf{Tattoo Removal Net (TRNet)}] is a U-net architecture \cite{skindeep} \cite{howard2018fastai} with spectral normalization and self-attention trained on our synthetic data (Sect.\ref{sec:generated_database}). The U-Net architecture is depicted in Fig.~\ref{fig:tnet_arch}; the encoder is based on ResNet34 and the decoder consist of four main blocks and utilize PixelShuffling \cite{Shi-Pixelshuffle-CVPR-2016}. The loss function is a combination of feature loss (perceptual loss) from \cite{Johnson-PerceptualLoss-ECCV-2016}, gram matrix style loss \cite{Gatys-ArtisticStyle-arxiv-2015}, and pixel (L1) loss. For the gram matrix loss and the feature loss, blocks from a pre-trained VGG-16 model is used \cite{Liu-VGG-ACPR-2015} \cite{howard2018fastai}.
\end{description}

\subsection{Quality Metrics}
\label{sec:metrics} 
To evaluate the quality of the different tattoo removal models, we use three different metrics commonly used in the literature: 

\begin{description}
\item[Peak signal-to-noise ratio (\textbf{PSNR})] is a measurement of error between an input and an output image and is calculated as follows:

$$\mathit{PSNR}(X, Y) = 20 \cdot log_{10}\ (\frac{\mathit{MAX}_{I}}{\sqrt{\mathit{MSE}(X, Y)}})$$

where $\mathit{MAX}_{I}$ is the theoretical maximum pixel value (\ie 255 for 8 bit channels) and $\mathit{MSE}(X, Y)$ is the mean squared error between the ground truth image $X$ and the inpainted image $Y$. The PSNR is measured in decibel and a higher value indicates better quality of the reconstructed image.
\end{description}

\begin{description}
\item[Mean Structural Similarity Index (\textbf{MSSIM})] as given in \cite{Wang-ImageQualityAssesment-SSIM-IEEE-2014}, is defined as follows:

$$\mathit{MSSIM}(X.Y) = \frac{1}{M} \sum_{i=1}^{M} \mathit{SSIM}(x_i, y_i)$$

where X and Y are the ground truth image and inpainted image, respectively, $M$ is the number of local windows in an image and $x_i$ and $y_i$ are the image content of the $i$'th local window. The SSIM over local window patches $(x,y)$ is defined as: 

$$ \mathit{SSIM}(x, y)  = \frac{(2 \mu_x \mu_y + C_1) (2 \sigma_{xy} + C_2)}{(\mu_x^2 + \mu_y^2 + C_1) + (\sigma_x^2 + \sigma_y^2 + C_2)}$$

where $\mu_x$ and $\mu_y$ are the mean values of the local window patches $x$ and $y$, respectively; $\sigma_x^2 + \sigma_y^2$ are their local variances and $\sigma_{xy}$ is the local covariance of $x$ and $y$. $C_1$ and $C_2$ are constants set based on the same parameter settings as Wang \etal \cite{Wang-ImageQualityAssesment-SSIM-IEEE-2014}, \ie $C_1 \approx 6.55, C_2 \approx 58.98$. MSSIM returns a value in the range of 0 to 1, where 1 means that X and Y are identical.
\end{description}

\begin{description}
\item[Visual Information Fidelity (\textbf{VIF})] is a full reference image quality assessment measurement proposed by Sheikh and Bovik in \cite{Sheikh-VIF-IEEE-2006}. VIF is derived from a statistical model for natural scenes as well as models for image distortion and the human visual system. $VIF(X,Y)$ returns a value in range of 0 to 1, where 1 indicates that the ground truth and inpainted images are identical. We use the pixel domain version as implemented in \cite{sewar}. 
\end{description}

\begin{figure}[!htbp]
    \begin{subfigure}{0.48\linewidth}
    \centering
        \includegraphics[width=0.6\textwidth]{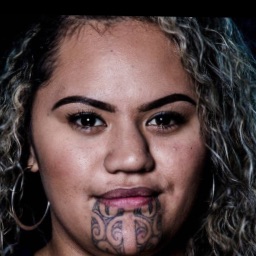}
        \caption{Portrait}
    \end{subfigure}
    \begin{subfigure}{0.48\linewidth}
    \centering
        \includegraphics[width=0.6\textwidth]{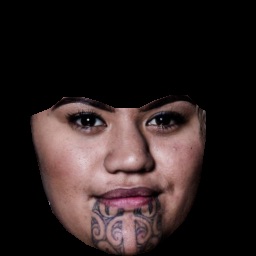}
        \caption{Inner}
    \end{subfigure}
    \caption{Examples of (a) a full portrait image where the entire face is visible and (b) a crop of the inner face region.}
    \label{fig:portrait_inner_examples}
\end{figure}

We estimate the different quality metrics both on portrait images, \ie where the entire face is visible and on the inner part of the face (corresponding to the area covered by the 68 dlib landmark points; see Fig.~\ref{fig:dlib_landmarks}) where we focus on only the area from the eyebrows to the chin; these regions are shown in Fig.~\ref{fig:portrait_inner_examples}.

\begin{figure*}[!htbp]
    \centering
    \includegraphics[width=0.75\linewidth]{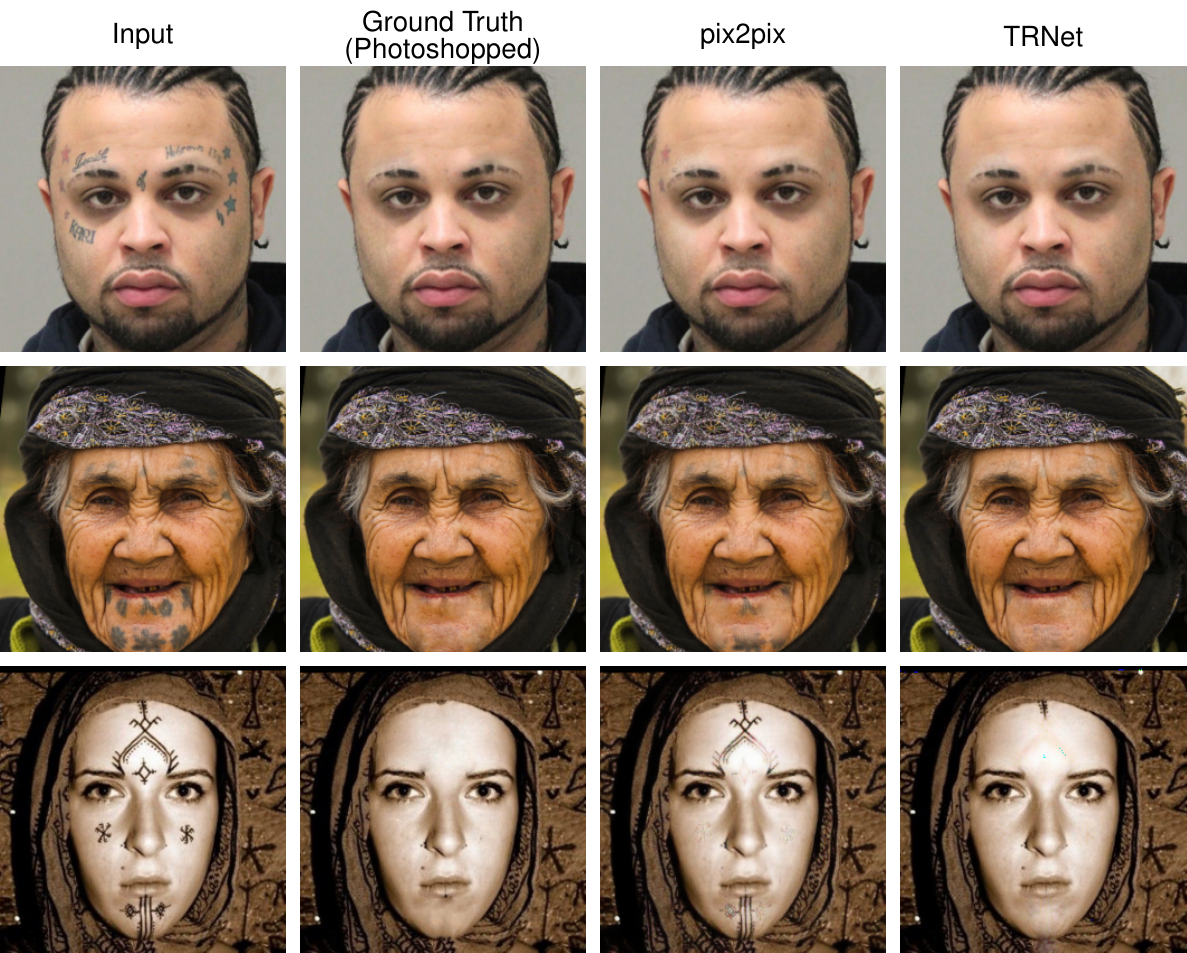}
    \caption{Examples of using deep-learning based algorithms for facial tattoo removal. Best viewed in electronic format (zoomed in).}
    \label{fig:removal_examples_quality_measurements}
\end{figure*}
\subsection{Removal Quality Results}
\label{sec:tattoo_removal_results}
We use a total of 41 facial images with tattoos where the tattoos have been manually removed using PhotoShop; we refer to these as our ground truth images. Examples of using the different deep learning-based methods for removing tattoos are given in Fig. \ref{fig:removal_examples_quality_measurements}. As seen, the best model (TRNet) is able to remove most tattoos with only a few artefacts, whereas the other models perform less well and, for some images, alter the face or fail to remove all tattoos accurately.

Different quality scores are reported in Tab.~\ref{tab:qual_score_removal} which shows that the TRNet model performs best in most scenarios especially when only looking at the inner part of the face.

\begin{table*}[htbp!]
\centering
\caption{Quality measurements of the reconstructed images compared to ground truth images where tattoos have been manually removed. "Tattooed" denotes the baseline case where the tattooed images are compared to the ground truth images.}
\label{tab:qual_score_removal}
\begin{adjustbox}{max width=\linewidth}
\begin{tabular}{@{\extracolsep{2pt}}lcccccc@{}} \toprule 
\multicolumn{1}{c}{} \multirow{2.5}{*}{\textbf{Scenario}} & \multicolumn{3}{c}{\textbf{Portrait}} & \multicolumn{3}{c}{\textbf{\begin{tabular}[c]{@{}c@{}}\textbf{Inner}\end{tabular}}} \\  \cmidrule{2-4} 
\cmidrule{5-7} 
 & \textbf{MSSIM} & \textbf{PSNR}  &  \textbf{VIF} & \textbf{MSSIM} &  \textbf{PSNR} & \textbf{VIF}  \\ \midrule
Tattooed & $0.947 \ (\pm 0.053)$ & $31.31 \ (\pm 5.04)$ & $ 0.884\ (\pm 0.093)$ & $0.974 \ (\pm 0.027)$ & $35.37 \ (\pm 6.63)$ & $0.879\ (\pm0.097)$ \\ 
pix2pix & $0.943 \ (\pm 0.043)$ & $33.24 \ (\pm 4.82)$ & $0.732\ (\pm0.081)$ & $0.978 \ (\pm 0.021)$ & $37.66 \ (\pm 5.39)$ & $0.779\ (\pm0.087)$ \\  
TRNet & $0.967\ (\pm0.034)$  & $36.22 \ (\pm6.00)$ & $0.883\ (\pm0.079)$ & $0.987\ (\pm 0.015)$ & $42.34 \ (\pm 6.74)$ & $0.891\ (\pm 0.083)$  \\   \bottomrule
\end{tabular}
\end{adjustbox}
\end{table*}

\begin{figure}[!htbp]
\centering
\begin{subfigure}[t]{.48\columnwidth}
\centering
\includegraphics[width=.48\columnwidth]{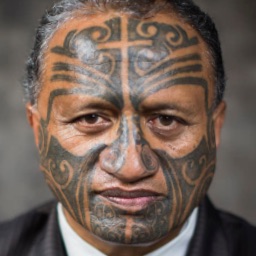}
\includegraphics[width=.48\columnwidth]{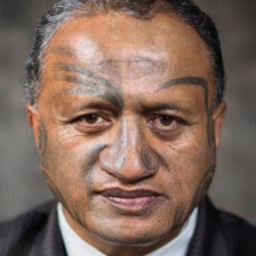}
\end{subfigure}\quad
\begin{subfigure}[t]{.48\columnwidth}
\centering
\includegraphics[width=.48\columnwidth]{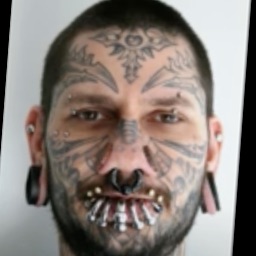}
\includegraphics[width=.48\columnwidth]{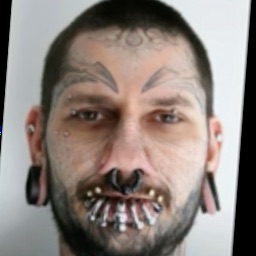}
\end{subfigure}
\begin{subfigure}[t]{.48\columnwidth}
\centering
\vspace{-6pt}
\includegraphics[width=.48\columnwidth]{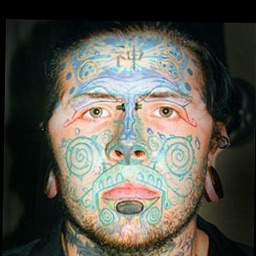}
\includegraphics[width=.48\columnwidth]{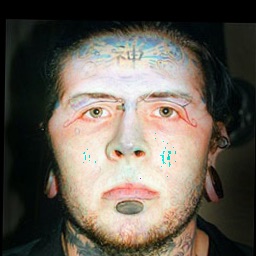}
\end{subfigure}\quad
\begin{subfigure}[t]{.48\columnwidth}
\centering
\vspace{-6pt}
\includegraphics[width=.48\columnwidth]{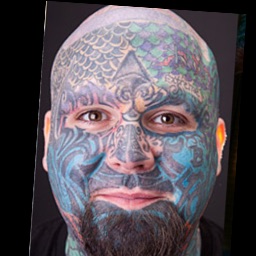}
\includegraphics[width=.48\columnwidth]{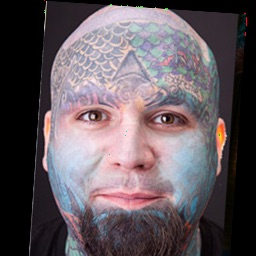}
\end{subfigure}
\caption{Facial images with extreme coverage of tattoos, which remain challenging for our tattoo removal approach. Before (left) and after (right) tattoo removal. }
\label{fig:extreme_examples}
\end{figure}

While the tattoo removal performs well in many scenarios, there are also some extreme cases where it does not work so well. Examples of removing large coverage of tattoos from facial images are shown in Fig.~\ref{fig:extreme_examples}. Depicted example images clearly show the limitations of the presented approach.

%% file: sections/results.tex
\section{Application to Face Recognition}
\label{sec:experiments}
In this section, we describe how tattoo removal can be integrated and used in a face recognition system. A face recognition system consists of several preprocessing modules such as face alignment and quality estimation. These modules help to minimise factors which are unimportant for face recognition and ensure that only images of sufficient quality are used during authentication. As part of the preprocessing, we propose to use the deep learning-based removal algorithms described in Sect.~\ref{sec:tattoo_removal}. While facial tattoos can be seen as distinctive and helpful in identifying individuals, tattoo removal is useful for face recognition in cases where only one of the face images in a comparison contains tattoos \cite{Ibsen-ImpactFacialTattoosPaintingsFaceRecognitionSystems-BMT-2021}. In our experiments, we trained the classifiers to remove facial tattoos from aligned images and as such will assume that our input images have already been aligned since our focus is on improving feature extraction and comparison. Note that the proposed tattoo removal method could also be retrained on unaligned images and placed before the detection module to improve detection accuracy.

\subsection{Experimental Setup}
In the following, we describe the database, the employed face recognition system, and metrics used to evaluate the biometric performance:

\begin{description}
\item[\textbf{Database}] for the evaluation, we use the publicly available database \textit{HDA Facial Tattoo and Painting Database}\footnote{\url{https://dasec.h-da.de/research/biometrics/hda-facial-tattoo-and-painting-database}}, which consists of 250 image pairs of individuals with and without real facial tattoos. The database was originally collected by Ibsen \etal in \cite{Ibsen-ImpactFacialTattoosPaintingsFaceRecognitionSystems-BMT-2021}. The images have all been aligned using the RetinaFace facial detector \cite{Deng-RetinaFace-CVPR-2020}. Examples of original image-pairs (before tattoo removal) are given in Fig.~\ref{fig:ibsen_tattoop_org_examples}. These pairs of images are used for evaluating the performance of a face recognition system. For evaluating the effect of tattoo removal, the models described in Sect.~\ref{sec:tattoo_removal_models} are employed on the facial images containing tattoos, whereafter the resulting images are used during the evaluation.

\begin{figure}[!htb]
\centering
\begin{subfigure}[t]{.48\columnwidth}
\centering
\includegraphics[width=.48\columnwidth]{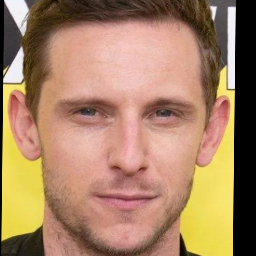}
\includegraphics[width=.48\columnwidth]{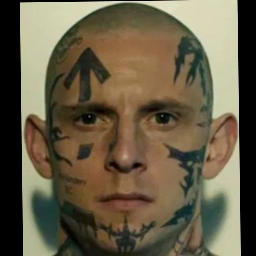}
\end{subfigure}\quad
\begin{subfigure}[t]{.48\columnwidth}
\centering
\includegraphics[width=.48\columnwidth]{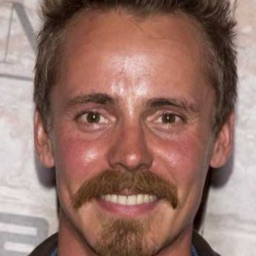}
\includegraphics[width=.48\columnwidth]{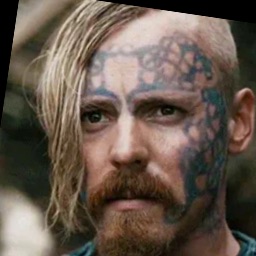}
\end{subfigure}
\caption{Examples of image-pairs in the HDA facial tattoo and painting database.}
\label{fig:ibsen_tattoop_org_examples}
\end{figure}
\end{description}

\begin{description}
\item[\textbf{Face recognition system}] to evaluate the applicability of tattoo removal for face recognition, we use the established ArcFace pre-trained model (LResNet100E-IR,ArcFace@ms1m-refine-v2) with the RetinaFace facial detector. 
\end{description}

\begin{description}
\item[\textbf{Recognition performance metrics}] the effect of removing facial tattoos is evaluated empirically \cite{ISO-IEC-19795-1-Framework-210216}. Specifically, we measure the FNMR at operationally relevant thresholds corresponding to a FMR of $0.1\%$ and $1\%$: 

\begin{itemize}
    \item \textbf{False Match Rate (FMR)}: the proportion of the completed biometric non-mated comparison trials that result in a false match.
    \item \textbf{False Non-Match Rate (FNMR)}: the proportion of the completed biometric mated comparison trials that result in a false non-match.
\end{itemize}

Additionally, we report the Equal Error Rate (EER), \ie the point where FNMR and FMR are equal. To show the distribution of comparison scores, boxplots are used. The comparison scores are computed between pairs of feature vectors using the Euclidean distance.
\end{description}

\subsection{Experimental Results}
\begin{figure}
    \centering
    \includegraphics[width=\linewidth]{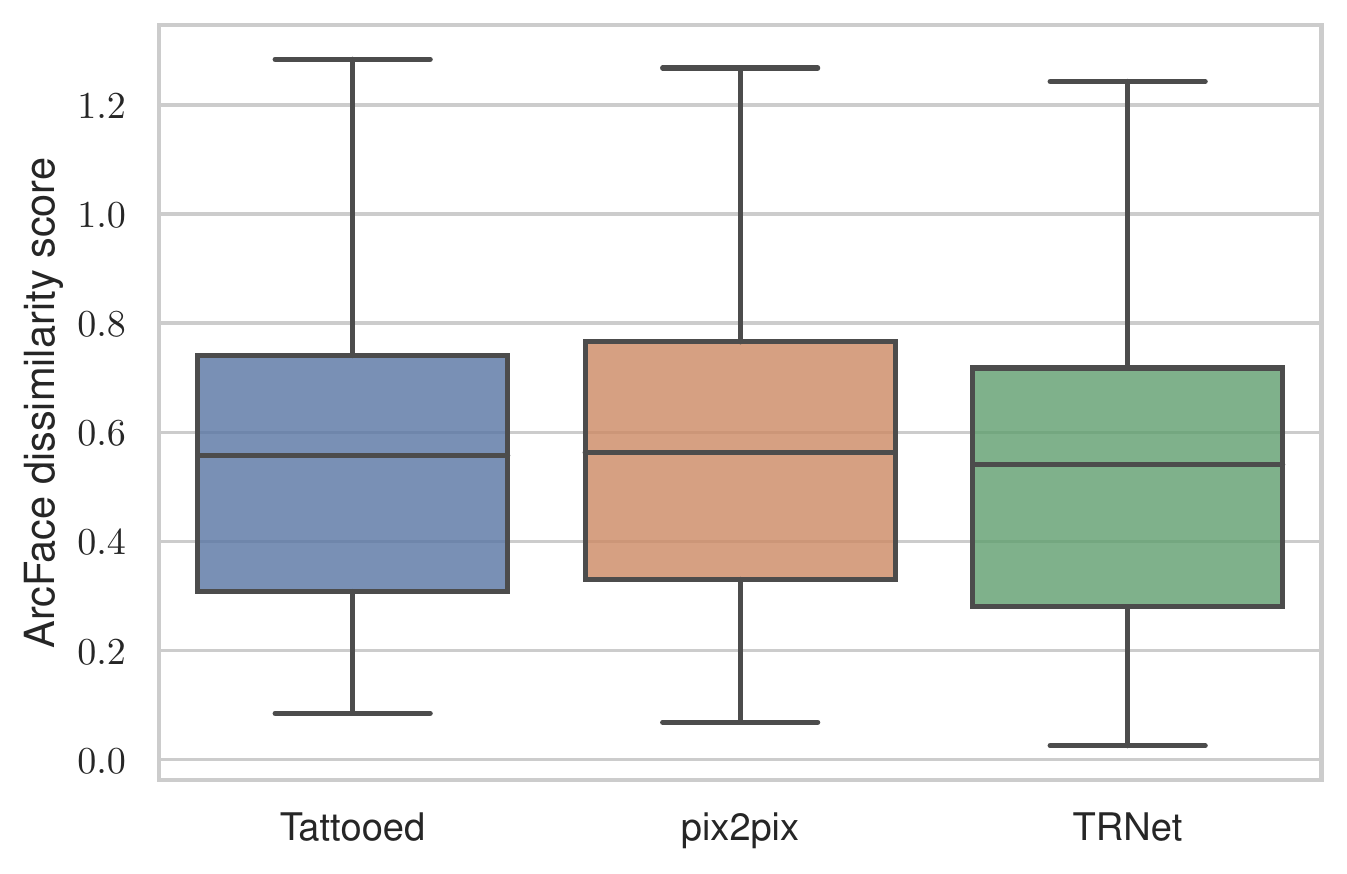}
    \caption{Boxplots showing the effect of tattoo removal on biometric comparison scores.}
    \label{fig:box_plots_dissimilarity}
\end{figure}

The effect of removing tattoos on the computed comparison scores is visualised in Fig.~\ref{fig:box_plots_dissimilarity}. As can be seen, the comparison scores are not significantly affected for the pix2pix model which only showed moderate capabilities of removing tattoos from facial images. However, for TRNet, which has been trained on the synthetic database, it is shown that the dissimilarity score on average gets lower, which indicates that the recognition performance might improve. 

\begin{table}[!htbp]
    \centering
\caption{Biometric performance results for ArcFace.}
   \begin{tabular}{@{\extracolsep{2pt}}lrrrrrr@{}} \toprule 
   \multirow{2.5}{*}{\textbf{Type}}  & \multirow{2.5}{*}{\textbf{EER\%}} & \multicolumn{2}{c}{\textbf{FNMR\%}} \\ \cmidrule{3-4} 
     & & \textbf{FMR$=$0.1\%}  & \textbf{FMR$=$1\%} \\ \midrule
        Tattooed &  $0.80$ & $1.20$ & $0.80$  \\
        pix2pix &  $0.80$ & $1.60$  & $0.80$\\
        TRNet &  $0.40$ & $1.20$ & $0.00$ \\
        \bottomrule
    \end{tabular}
\label{tab:biometric_performance_scores}
\end{table}

Table \ref{tab:biometric_performance_scores} shows the biometric performance scores calculated on the tattooed images and the inpainted facial images for the different used models. The scores indicate that realistic removal of tattoos (TRNet) might improve face recognition performance since we can observe that, compared to the baseline (tattooed), the EER is halved, and the FNMR at an FMR of $1\%$ is reduced to $0\%$. The results indicate that a tattoo removal module can be integrated into the processing chain of a face recognition system and help make it more robust towards facial tattoos. 

%% file: sections/conclusion.tex
\section{Summary}
\label{sec:conclusion}
In this paper, we proposed an automatic approach for blending tattoos onto facial images and showed that it is possible to use synthetic data to train a deep learning-based facial tattoo removal algorithm, thereby enhancing the performance of a state-of-the-art face recognition system. To create a facial image with tattoos, the face is first divided into face regions using landmark detection whereafter tattoo placements can be found. Subsequently, deep reconstruction maps and cut-out maps can be estimated from the input image. Thereafter, the information is combined to realistically blend tattoos to the facial image. Using this approach, we created a large database of facial images with tattoos and used it to train a deep learning-based algorithm for removing tattoos. Experimental results show a high quality of the tattoo removal. To further show the feasibility of the reconstruction, we evaluated the effect of removing facial tattoos on a state-of-the-art face recognition system and found that it can improve automated face recognition performance. 

%% file: sections/acknowledgement.tex
\section*{Acknowledgements}
This research work has been funded by the German Federal Ministry of Education and Research and the Hessian Ministry of Higher Education, Research, Science and the Arts within their joint support of the National Research Center for Applied Cybersecurity ATHENE and the European Union’s Horizon 2020 research and innovation programme under the Marie Skłodowska-Curie grant agreement No. 860813 - TReSPAsS-ETN.